# Learning from Noisy Label Distributions

Yuya Yoshikawa

Software Technology and Artificial Intelligence Research Laboratory (STAIR Lab),
Chiba Institute of Technology, Japan.
yoshikawa@stair.center

**Abstract.** In this paper, we consider a novel machine learning problem, that is, learning a classifier from noisy label distributions. In this problem, each instance with a feature vector belongs to at least one group. Then, instead of the true label of each instance, we observe the label distribution of the instances associated with a group, where the label distribution is distorted by an unknown noise. Our goals are to (1) estimate the true label of each instance, and (2) learn a classifier that predicts the true label of a new instance. We propose a probabilistic model that considers true label distributions of groups and parameters that represent the noise as hidden variables. The model can be learned based on a variational Bayesian method. In numerical experiments, we show that the proposed model outperforms existing methods in terms of the estimation of the true labels of instances.

## 1   Introduction

In this paper, we consider a novel machine learning problem, that is, learning a classifier from noisy label distributions. Figure 1 illustrates the assumptions for this problem. There are $N$ groups and $U$ instances. Each instance has a feature vector and true single label. Each group consists of a subset of all instances. Then, for each group, the *true label distribution*, that is, the distribution of the true labels of the instances associated with the group, can be calculated. However, the true labels and true label distribution cannot be observed. Instead, we observe the *noisy label distribution*, that is, the label distribution such that the true label distribution is distorted by an unknown noise. Our goals are to (1) estimate the true label of each instance, and (2) learn a classifier that predicts the true label of a new instance.

We propose a generative probabilistic model that considers the true label distributions of groups and parameters that represent the noise as hidden variables. The model can be learned based on a variational Bayesian method [1]. In numerical experiments, we show that, using a synthetic dataset generated based on the problem, the proposed model outperforms existing methods in terms of the estimation of the true labels of instances.

As a particular example of this problem, we consider the demographic estimation of individuals, which is the estimation of gender, age, occupation, race, and living place using individuals' features [9,2,6]. In social networking services (SNS) on the web, marketing and developing advertisement delivery systems are conducted using the demographic information of users. However, because such demographic information cannot be obtained in many cases, its estimation is required. A typical approach to demographic estimation using machine learning is to first annotate the demographic information for individuals and then learn a classifier that predicts the demographics for unknown individuals using the annotated demographic information. However, it is difficult and expensive to annotate the information manually.

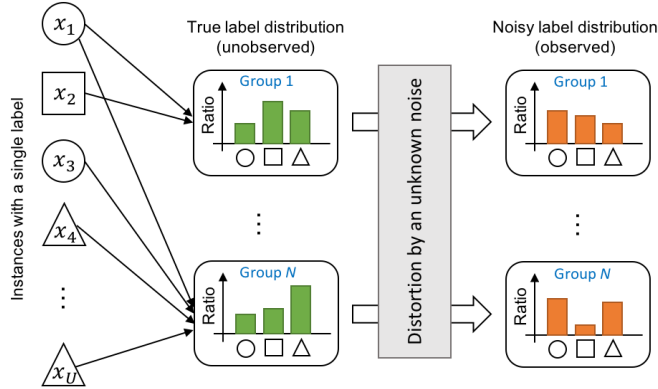

**Fig. 1.** Assumptions for learning a classifier from noisy label distributions. The shape of each instance indicates its label. The edge connecting an instance to a group indicates that the instance belongs to the group.

Instead, using the machine learning approach, we can use the proposed model for demographic estimation. For the case of demographic estimation, each instance corresponds to a user on SNS, and its feature vector and label are the content created by the user and the demographic label of the user, respectively. Then, we use corporate accounts on SNS as groups and regard the users who follow the corporate accounts as members of the groups. Because the demographic labels of the users are unknown, we cannot also observe the true label distributions of the groups. Instead, we use the demographic distributions of visitors to corporate websites, which can be obtained easily and cheaply from audience measurement services, such as Quantcast[1]. Because the demographic distributions of websites differ from those of SNS, we use the distributions as noisy label distributions in our problem. Finally, by learning the proposed model, it is expected that the demographics of the users in SNS can be estimated without annotating demographic information.

## 2   Related Work

To the best of our knowledge, there is no study that addresses the problem contained in this paper. However, this problem may be considered to be similar to or a type of multiple instance learning (MIL) [5,7]. In standard MIL, the sets that consist of pairs of instances and their binary labels are given. Each of the sets is labeled positive if at least one positive label is included in the set, and negative otherwise. Then, the goal of MIL is to learn a classifier that predicts whether a new set that consists only of instances is positive or negative. There are two main differences between our problem and MIL. First, our problem assumes that the label of each instance is unobserved. Second, the goal of our problem is to learn a classifier that predicts a label of a newly given instance rather than a set of instances.

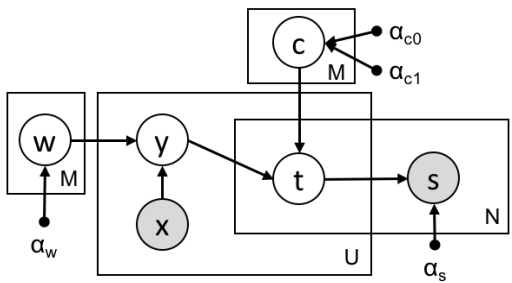

**Fig. 2.** Graphical representation of the proposed model. The white and the gray circles represent hidden and observed variables, respectively. The edges represent the dependency between the variables. The dots are hyper-parameters determined manually before training.

Note that our problem is inspired by the demographic estimation method of Culotta et al. [4,3]. They considered the situation described in the last paragraph of Section 1. Then, they proposed learning a demographic classifier of SNS users directly using the demographic distributions of website visitors. However, their method implicitly assumes that the demographic distributions of SNS users and website visitors are identical.

In this study, we formalize the problem of Culotta et al. as a general machine learning problem. Then, the proposed model can capture the difference between two distributions, such as the demographic distributions of SNS users and website visitors, by modeling the process by which they are distorted by an unknown noise.

## 3   Proposed Method

In this section, we propose a probabilistic model to address the problem described in the first paragraph of Section 1. First, we explain the formulation of the proposed model. Then, we explain how to learn the proposed model based on a variational Bayesian method.

### 3.1   Model Formulation

In the problem, there are $N$ groups and $U$ instances. Each instance has a $D$-dimensional feature vector and true label. Let $\mathbf{x}_u \in \mathbb{R}^D$ be the feature vector of the $u$th instance and $y_u \in \{1, 2, \cdots, M\}$ be the true label of the $u$th instance, where $M$ is the number of classes. For convenience, we define a set of all feature vectors and set of all true labels as $\mathcal{X} = \{\mathbf{x}_u\}_{u=1}^U$ and $\mathcal{Y} = \{y_u\}_{u=1}^U$, respectively. Each group consists of a subset of all instances. We denote the set of instances associated with the $i$th group by $G_i \subseteq \{u\}_{u=1}^U$. Then, for each group, the *true label distribution*, that is, the distribution of the true label of the instances associated with the group, can be calculated. In particular, we denote the true label distribution of the $i$th group by $\mathbf{z}_i \in \mathbb{R}^M$, where $\mathbf{z}_i = \frac{1}{|G_i|} \sum_{u \in G_i} \mathrm{vec}(y_u)$ and $\sum_{m=1}^M z_{im} = 1$, where $\mathrm{vec}(y_u)$ returns a vector whose $y_u$th element is one and other elements are zero. Note that, the true label of each instance and true label distribution of each group are unobserved. Instead, we observe the *noisy label distribution*, that is, a label distribution such that the true label distribution is distorted by an unknown noise. We denote the noisy label distribution of the $i$th group by $\mathbf{s}_i \in \mathbb{R}^M$, where $\sum_{m=1}^M s_{im} = 1$.

Figure 2 illustrates the graphical representation of the proposed model. The generative process of the proposed model is as follows:

1. For each class $m = 1, 2, \cdots, M$:
   (a) Draw weight vector $\mathbf{w}_m \sim \mathcal{N}(\mathbf{0}, \alpha_w^{-1}\mathbf{I}_D)$.
   (b) Draw confusion vector $\mathbf{c}_m \sim \mathrm{Dir}(\beta_m)$.
2. For each instance $u = 1, 2, \cdots, U$:
   (a) Draw true label $y_u \sim \mathrm{Softmax}([\mathbf{w}_m^\top \mathbf{x}_u]_{m=1}^M)$.
   (b) For each group $i = 1, 2, \cdots, N$:
      i. Draw group-dependent label $t_i \sim \mathrm{Cat}(\mathbf{c}_{y_u})$.
3. For each group $i = 1, 2, \cdots, N$:
   (a) Draw noisy label distribution $\mathbf{s}_i \sim \mathcal{N}(\mathbf{t}_i, \alpha_s^{-1}\mathbf{I}_M)$,
       where $\mathbf{t}_i = \frac{1}{|G_i|} \sum_{u \in G_i} \mathrm{vec}(t_{iu})$.

In the proposed model, the true label $y_u$ of each instance $u$ is generated from the distribution calculated based on the inner product of weight vector $\mathbf{w}_m$ and feature vector $\mathbf{x}_u$. Weight vector $\mathbf{w}_m$ for each class $m$ is generated from an isotropic Gaussian with zero mean and covariance matrix $\alpha_w^{-1}\mathbf{I}_D$, where $\mathbf{I}_D$ is a $D$-dimensional identity matrix. The distribution is normalized using a softmax function. Although this is the same idea as multi-class logistic regression, in the proposed model, the true label is unobserved and considered as a hidden variable.

The true label distribution (unobserved) and noisy label distribution (observed) are different because of the distortion by an unknown noise. To capture such a phenomenon, the proposed model has confusion vector $\mathbf{c}_m \in \mathbb{R}_+^M$ for each class $m$, where the $l$th element of $\mathbf{c}_m$ represents the probability with which class $m$ is changed to class $l$ by an unknown noise. We assume that confusion vector $\mathbf{c}_m$ for each class $m$ is generated from the Dirichlet distribution, with parameter $\beta_m \in \mathbb{R}_+^M$. To incorporate the magnitude of the noise as prior knowledge into the proposed model, we parameterize $\beta_m$ as follows:

$$\beta_{ml} = \begin{cases} \alpha_{c_0} & (m \neq l) \\ \alpha_{c_1} & (m = l). \end{cases} \tag{1}$$

When a large magnitude of noise is expected, we set $\alpha_{c_0}$ to a larger value than $\alpha_{c_1}$.

Then, the noisy label distribution of each group is generated. In particular, for each instance $u \in G_i$ associated with the $i$th group, group-dependent label $t_{iu} \in \{1, 2, \cdots, M\}$ is generated from the categorical distribution, with parameter $\mathbf{c}_{y_u}$. We define $\mathbf{t}_i = \frac{1}{|G_i|} \sum_{u \in G_i} \text{vec}(t_{iu})$. Note that, because the instance can belong to multiple groups, the group-dependent labels of the instance may vary according to the groups. Finally, the noisy label distribution of the $i$th group is generated from a Gaussian distribution, with $\mathbf{t}_i$ as a mean vector and $\alpha_s^{-1} \mathbf{I}_M$ as a covariance matrix.

### 3.2   Inference Based on Variational Bayesian Method

We introduce the inference method of the proposed model based on a variational Bayesian method [1].

First, the logarithm of the marginal posterior of weight matrix $\mathbf{W}$ and confusion matrix $\mathbf{C}$ is given by

$$\log p(\mathbf{W}, \mathbf{C}|\mathbf{X}, \mathbf{S}, \alpha) \propto \log \sum_{\mathbf{T}, \mathbf{Y}} p(\mathbf{S}|\mathbf{T}, \alpha_s) p(\mathbf{T}|\mathbf{Y}, \mathbf{C}) p(\mathbf{Y}|\mathbf{W}, \mathbf{X})$$
$$+ \log p(\mathbf{W}|\alpha_w) + \log p(\mathbf{C}|\alpha_{c_0}, \alpha_{c_1}). \quad (2)$$

According to the generative process, the factors in (2) are defined as follows:

$$p(\mathbf{S}|\mathbf{T}, \alpha_s) = \prod_{i=1}^{N} \mathcal{N}(\mathbf{s}_i | \mathbf{t}_i, \alpha_s^{-1} \mathbf{I}_M), \quad (3)$$

$$p(\mathbf{T}|\mathbf{Y}, \mathbf{C}) = \prod_{i=1}^{N} \prod_{u \in E_i} c_{y_u, t_{iu}}, \quad (4)$$

$$p(\mathbf{Y}|\mathbf{W}, \mathbf{X}) = \prod_{u=1}^{U} \frac{\exp(\mathbf{w}_{y_u}^\top \mathbf{x}_u)}{\sum_{m=1}^{} \exp(\mathbf{w}_m^\top \mathbf{x}_u)}, \quad (5)$$

$$p(\mathbf{W}|\alpha_w) = \prod_{m=1}^{M} \mathcal{N}(\mathbf{w}_m | \mathbf{0}, \alpha_w^{-1} \mathbf{I}_D), \quad (6)$$

$$p(\mathbf{C}|\alpha_{c_0}, \alpha_{c_1}) = \prod_{m=1}^{M} \frac{\Gamma\left(\sum_{l=1}^{M} \beta_{ml}\right)}{\prod_{l=1}^{M} \Gamma(\beta_{ml})} \prod_{l=1}^{M} c_{ml}^{\beta_{ml}-1}. \quad (7)$$

The goal of inference is to obtain $\mathbf{W}$ and $\mathbf{C}$ such that (2) is maximized. However, considering all possible combinations of $\mathbf{T}$ and $\mathbf{Y}$ is impossible in terms of the complexity of computational time. Therefore, we derive the following variational lower bound $\mathcal{L}(\Theta)$ for (2) according to Jensen's inequality:

$$\log p(\mathbf{W}, \mathbf{C}|\mathbf{X}, \mathbf{S}, \alpha) \geq \sum_{\mathbf{T}, \mathbf{Y}} q(\mathbf{T}, \mathbf{Y}) \log \frac{p(\mathbf{S}|\mathbf{T}, \alpha_s) p(\mathbf{T}|\mathbf{Y}, \mathbf{C}) p(\mathbf{Y}|\mathbf{W}, \mathbf{X})}{q(\mathbf{T}, \mathbf{Y})}$$
$$+ \log p(\mathbf{W}|\alpha_w) + \log p(\mathbf{C}|\alpha_{c_0}, \alpha_{c_1})$$
$$= \mathcal{L}(\Theta). \quad (8)$$

Then, we optimize the variational lower bound (8) instead. For convenience, we define $\Theta = \{\mathbf{W}, \mathbf{C}, \zeta, \eta\}$. With variational distribution $q(\mathbf{T}, \mathbf{Y})$, we assume the following factorization: $q(\mathbf{T}, \mathbf{Y}) = q(\mathbf{Y}|\zeta)q(\mathbf{T}|\eta)$, where

$$q(\mathbf{Y}|\zeta) = \prod_{u=1}^{U} \zeta_{uy_u}, \quad q(\mathbf{T}|\eta) = \prod_{i=1}^{N} \prod_{u=1}^{U} \eta_{iut_{iu}}. \tag{9}$$

Next, we derive the update rules for $\mathbf{W}$, $\mathbf{C}$, $\zeta$, and $\eta$.

*Update* $\mathbf{W}$. Because $\mathbf{W}$ cannot be updated in closed form, we calculate the gradient with respect to $\mathbf{W}$ as follows:

$$\frac{\partial \mathcal{L}(\Theta)}{\partial \mathbf{w}_m} = \sum_{u=1}^{U} \left( \zeta_{um} - \frac{\exp(\mathbf{w}_m^\top \mathbf{x}_u)}{\sum_{l=1}^{M} \exp(\mathbf{w}_l^\top \mathbf{x}_u)} \sum_{l=1}^{M} \zeta_{ul} \right) \mathbf{x}_u - \alpha_w \mathbf{w}_m. \tag{10}$$

Then we update $\mathbf{W}$ using a gradient-based optimization method, such as the quasi-Newton method.

*Update* $\mathbf{C}$. Because of the constraint that $\sum_{l=1}^{M} c_{ml} = 1$ for each class $m$, we derive the update rule for $\mathbf{C}$ according to the Lagrange multiplier method as follows:

$$c_{ml} = \frac{\sum_{i=1}^{N} \sum_{u=1}^{U} \zeta_{um} \eta_{ium} + \beta_{ml} - 1}{\sum_{m'=1}^{M} \sum_{i=1}^{N} \sum_{u=1}^{U} \zeta_{um'} \eta_{ium'} + \sum_{l'=1}^{M} \beta_{ml'} - M}. \tag{11}$$

*Update* $\zeta$. Because of the constraint that $\sum_{m=1}^{M} \zeta_{um} = 1$ for each instance $u$, we derive the update rule for $\zeta$ according to the Lagrange multiplier method as follows:

$$\zeta_{um} \propto \exp \left\{ \sum_{i=1}^{N} \sum_{l=1}^{M} \eta_{iul} \log c_{ml} + a_{um} - \log \sum_{m'=1}^{M} \exp(a_{um'}) \right\}, \tag{12}$$

where $a_{um} = \mathbf{x}_u^\top \mathbf{w}_m$. After calculating (12), the values are normalized such that $\sum_{m=1}^{M} \zeta_{um} = 1$.

*Update* $\eta$. Because of the constraint that $\sum_{m=1}^{M} \eta_{ium} = 1$ for the pair of each group $i$ and each instance $u$, we derive the update rule for $\eta$ according to the Lagrange multiplier method. However, because the update rule cannot be calculated in closed form, we derive the gradients with respect to $\eta_{ium}$ and multiplier parameter $\lambda_{iu}$ as follows:

$$\frac{\partial \mathcal{L}(\Theta)}{\partial \eta_{ium}} = -\frac{\alpha_s}{|G_i|} \left( s_{im} - \mathbb{E}[\mathbf{t}_i]_m \right) + \sum_{l=1}^{M} \zeta_{ul} \log c_{lm} - \log \eta_{ium} - 1 + \lambda_{iu}, \tag{13}$$

$$\frac{\partial \mathcal{L}(\Theta)}{\partial \lambda_{iu}} = \sum_{m=1}^{M} \eta_{ium} - 1. \tag{14}$$

and then we alternatively update the parameters using these gradients, where $\mathbb{E}[\mathbf{t}_i]_m = \frac{1}{|G_i|} \sum_{u \in G_i} \eta_{ium}$.

We continue to update the parameters $\mathbf{W}$, $\mathbf{C}$, $\zeta$, and $\eta$ sequentially until the value of (8) converges. The hyper-parameters $\alpha$ are determined by cross-validation.

**Table 1.** Accuracy of true label estimation on a synthetic dataset.

|          | $\alpha_{c_1} = 1$ | $\alpha_{c_1} = 10$ | $\alpha_{c_1} = 100$ |
|----------|:------:|:------:|:------:|
| Proposed | **0.43** | **0.52** | **0.45** |
| MTEN [4] | 0.32 | 0.51 | 0.31 |
| Ridge [4] | 0.24 | 0.48 | 0.24 |

## 4    Experiments

To confirm the effectiveness of the proposed model in the scenario shown in Figure 1, we performed numerical experiments on a synthetic dataset.

We considered a four-class classification problem. The synthetic dataset was generated according to the generative process of the proposed model described in Section 3.1. Note the following:

  – The feature vector of each instance was generated from an isotropic Gaussian with zero mean and a variance equal to one.
  – We set the number of instances to $U = 100$ and the number of groups to $N = 1,000$, and each group consisted of 30 randomly chosen instances.
  – We defined $\beta$ according to (1), where we set $\alpha_{c_0} = 1$ and $\alpha_{c_1} \in \{1, 10, 100\}$.

For comparison, we used two methods proposed by Culotta et al. [4]. These methods learn a regression function that predicts the values of the noisy label distribution of each group from the feature vector of each instance associated with the group. After learning, they calculate the label distribution of a newly given instance, and then output the label with the highest value of the label distribution as a prediction result. Culotta et al. used a multi-task elastic net (MTEN) that captured the relationship among each dimension of the distribution, and a ridge regression that predicted the value of each dimension of the distribution independently. For MTEN and ridge regression, we used the implementation of scikit-learn [8] in the same way as Culotta et al.

Table 1 shows the accuracy of the true label estimation on three synthetic datasets with different $\alpha_{c_1}$. When $\alpha_{c_1}$ was small, the difference between the true label distribution and the noisy label distribution was large. As a result, the accuracy of the existing method, for example, ridge regression, was much the same as a random choice, that is, $1/4 = 0.25$. By contrast, because the proposed model could capture the difference by learning confusion matrix $\mathbf{C}$, it could estimate the true labels accurately. Moreover, when $\alpha_{c_1}$ was large, the proposed model consistently outperformed existing methods.

## 5    Conclusion

We considered a novel machine learning problem, that is, learning a classifier from noisy label distributions. To address this problem, we proposed a generative probabilistic model, which can be inferred based on a variational Bayesian method. In numerical experiments, we showed that, the proposed model outperformed existing methods in terms of the estimation of the true labels of instances.

In future work, we will confirm the effectiveness of the proposed model in demographic estimation using real datasets provided by Culotta et al. [3]. Additionally, because the proposed model only captured a single pattern of noise, we will extend the proposed model so that it captures multiple patterns of noise.

## References

1. Bishop, C.M.: Pattern Recognition and Machine Learning. Springer (2006)
2. Cheng, Z., Caverlee, J., Lee, K.: You Are Where You Tweet : A Content-Based Approach to Geo-locating Twitter Users. In: Proceedings of the 19th ACM International Conference on Information and Knowledge Management. pp. 759–768 (2010)
3. Culotta, A., Kumar, N.R., Cutler, J.: Predicting Twitter User Demographics Using Distant Supervision from Website Traffic Data. Journal of Artificial Intelligence Research 1 (2016)
4. Culotta, A., Ravi, N.K., Cutler, J.: Predicting the Demographics of Twitter Users from Website Traffic Data. In: Proceedings of the Twenty-Ninth AAAI Conference on Artificial Intelligence. pp. 72–78 (2015)
5. Dietterich, T.G., Lathrop, R.H., Lozano-Pérez, T.: Solving the Multiple Instance Problem with Axis-Parallel Rectangles. Artificial Intelligence 89, 31–71 (1997)
6. Li, J., Ritter, A., Hovy, E.: Weakly Supervised User Profile Extraction from Twitter. In: Association of Computational Linguistics. pp. 165–174 (2014)
7. Maron, O., Lozano-Pérez, T.: A Framework for Multiple-Instance Learning. In: Advances in Neural Information Processing. pp. 570–576 (1997)
8. Pedregosa, F., Varoquaux, G., Gramfort, A., Michel, V., Thirion, B., Grisel, O., Blondel, M., Prettenhofer, P., Weiss, R., Dubourg, V., Vanderplas, J., Passos, A., Cournapeau, D., Brucher, M., Perrot, M., Duchesnay, É.: Scikit-learn: Machine Learning in Python. Journal of Machine Learning Research 12(Oct), 2825–2830 (2011)
9. Rao, D., Yarowsky, D., Shreevats, A., Gupta, M.: Classifying Latent User Attributes in Twitter. In: Proceedings of the 2nd International Workshop on Search and Mining User-Generated Contents (2010)